\documentclass[11pt]{article}

\usepackage[final]{acl}

 \usepackage{microtype}
 \usepackage{kotex}

\usepackage{amsmath}
\usepackage{amssymb}

\usepackage{booktabs}
\usepackage{multirow}
\usepackage[table]{xcolor}
\usepackage{graphicx}
\usepackage{caption}
\usepackage{adjustbox}
\usepackage{makecell}
\usepackage{float}
\usepackage{tabularx}
\usepackage[most]{tcolorbox}
\usepackage[english,bidi=default]{babel} 
\babelprovide[import]{hindi}
\babelprovide[import]{arabic}

\usepackage{tabularx}
\usepackage{booktabs}
\usepackage{multirow}
\usepackage[table]{xcolor}

\newlength{\MarsTableWidth}

\makeatletter
\newcommand{\FullRowColor}[1]{%
  \makebox[0pt][l]{%
    \kern-\tabcolsep
    \color{#1}%
    \rule[-\dp\@arstrutbox]
         {\MarsTableWidth}
         {\dimexpr\ht\@arstrutbox+\dp\@arstrutbox\relax}%
  }%
}
\makeatother

\title{
\textbf{MARS}: What Retrieval Signals Are Hidden in Multimodal Large Language Models for Text-Video Retrieval?
}

\author{
Uicheol Jung,\quad
Juyoung Hong,\quad
Geuntaek Lim,\quad
Yukyung Choi \\[0.45em]
Sejong University, Seoul, Republic of Korea \\[0.35em]
\texttt{\{\href{mailto:ucjung@rcv.sejong.ac.kr}{\textcolor{black}{ucjung}},\href{mailto:ykchoi@rcv.sejong.ac.kr}{\textcolor{black}{ykchoi}}\}@rcv.sejong.ac.kr}
}
\begin{document}
\maketitle

\begin{abstract}
Text-video retrieval requires representations that can distinguish videos with similar scenes, actions, and temporal patterns. 
Recent multimodal large language models have been adapted as embedding models, but they often represent each input using a single token from the final layer. 
This can compress diverse video-text cues into a single vector and limit fine-grained retrieval.
To address this limitation, we propose MARS, a multi-layer and multi-slot embedding framework for text-video retrieval. 
MARS constructs multiple adaptive representation slots by combining hidden states from different decoder layers, compares corresponding text and video slots, and aggregates their similarities for retrieval. 
To better handle confusing candidates, we further introduce a hard-negative-aware slot specialization objective that encourages the slots to capture discriminative matching cues. 
Experiments on four text-video retrieval benchmarks show that MARS achieves state-of-the-art results in both direct similarity-based retrieval and reranking settings.
Ablation studies and analyses demonstrate that multi-layer fusion, multiple slots, and hard-negative-aware slot specialization provide complementary gains. Code is available at \url{https://github.com/sejong-rcv/MARS}.
\end{abstract}

\section{Introduction}
Text-video retrieval aims to align sentences and videos in a shared representation space based on their semantic similarity~\citep{miech2019howto100m,dong2022dual,gabeur2020multi,bain2021frozen}.
With the rapid growth of video data, this task has become important for retrieving relevant video content through natural language queries.

Recently, MLLMs have attracted increasing attention as general-purpose embedding models~\citep{jiang2024scaling,jiang2024e5v,jiang2025vlm2vec,zhang2025gme,liu2025lamra}.
Rather than generating text, these methods obtain embeddings from the hidden state of a specific token (e.g., the final or \(\langle\mathrm{EOS}\rangle\) token) given a prompted input.
For example, E5-V adopts an \texttt{in one word} prompting strategy inspired by PromptEOL~\citep{jiang2024scaling}, such as \texttt{<image> Summary above image in one word:}, to obtain a unified multimodal representation~\citep{jiang2024e5v}.
These studies show that MLLMs can be used as representation models for multimodal inputs.

However, adapting MLLM-based methods to text-video retrieval remains underexplored.
While extracting a single token representation from the final layer is effective for general-purpose tasks~\citep{jiang2024e5v,zhang2025gme}, it creates a severe information bottleneck for complex video content.
Because videos inherently contain multi-granular and compositional cues (e.g., objects, actions, and temporal dynamics), retrieval embeddings must preserve fine-grained semantic information to differentiate visually similar scenes.
This limitation motivates us to rethink embedding extraction from two perspectives: leveraging distributed information across multiple decoder layers, and preserving complementary cues without collapsing them into a single vector.

\textbf{First}, although final-layer representations are widely used, recent studies indicate they are not always optimal for downstream embedding tasks~\citep{skean2025layer,tang2024pooling}. Useful embedding signals are often distributed across multiple decoder layers rather than concentrated at the end. Accordingly, text-video retrieval can benefit from aggregating this multi-layer evidence instead of relying solely on the final output.

\textbf{Second}, extracted cues should be preserved rather than collapsed into a single vector. Prior work on fine-grained retrieval demonstrates that global representations alone struggle to capture detailed matching signals like objects and actions~\citep{chen2020fine,ma2022xclip}. Forcing these diverse cues into a single representation inevitably entangles them, obscuring critical semantic details and degrading performance on similar candidates. Therefore, to maintain fine-grained discriminability, these cues must be preserved and compared separately before computing the final retrieval.

To address these limitations, we propose \textbf{MARS} (\textbf{M}ulti-layer \textbf{A}daptive \textbf{R}epresentation \textbf{S}lots).
Rather than extracting an embedding from a single layer, MARS constructs each slot by fusing hidden states from multiple decoder layers with slot-specific layer weights.
This allows different slots to emphasize different layer-wise evidence.
Instead of compressing all information into one global vector, MARS preserves complementary cues through multiple adaptive representation slots and compares corresponding slots before aggregating their similarities.
We further introduce a hard-negative-aware slot specialization loss and a diversity loss to encourage the slots to capture distinct matching signals.
As a result, MARS can better distinguish candidates that differ in fine-grained semantic details.

To summarize, our contributions are as follows:


\begin{itemize}
\item We revisit embedding extraction for MLLM-based text-video retrieval and propose MARS, which constructs adaptive representation slots by fusing layer-wise representations.

\item We validate MARS on four benchmarks, achieving state-of-the-art results in both direct retrieval and reranking settings.

\item We provide in-depth analyses to elucidate how multi-layer evidence aggregation and complementary slot-wise matching contribute to the performance gains of MARS.
\end{itemize}

\section{Related Work}
\label{sec:related_work}

\subsection{Text-Video Retrieval}
A major line of work extends CLIP-based representations to the video domain through frame aggregation, temporal modeling, and fine-grained alignment~\citep{radford2021learning,luo2022clip4clip,xue2023clipvip,ma2022xclip,wu2023cap4video,wang2023ucofia,shen2025tempme,jung2026tame}.
In parallel, video foundation models have improved text-video retrieval through large-scale pre-training~\citep{li2023umt,wang2022internvideo,wang2024internvideo2}.
However, these gains often rely on heavy cross-modal decoders or reranking~\citep{li2023umt,wang2024internvideo2,ko2025blim}.
In contrast, our work follows an efficient dual-encoder retrieval structure, where videos and texts are encoded independently and ranked directly by representation similarity without additional pairwise scoring.


\subsection{MLLMs as Multimodal Embedders}
\label{sec:related_mllm_embedders}
Recent studies have adapted MLLMs into multimodal embedders through prompt-based extraction~\citep{jiang2024e5v,liu2025lamra} and contrastive or data-centric training~\citep{jiang2025vlm2vec,lin2025mmembed,zhang2025gme,chen2025mme5,zhou2025megapairs,meng2026vlm2vecv2}.
Despite methodological differences, these approaches predominantly derive the embedding from a single token (e.g., the final or $\langle\mathrm{EOS}\rangle$ token).
This single-token extraction fails to fully exploit the multi-layered representations within MLLMs, inherently limiting their ability to capture fine-grained cross-modal correspondences.
Instead of collapsing information into a single token from a single layer, our work constructs multiple layer-fused adaptive representation slots, preserving detailed semantic cues for accurate slot-wise matching.


\section{Method}
\label{sec:method}

\begin{figure*}[t]
    \centering
    \includegraphics[width=1\textwidth]{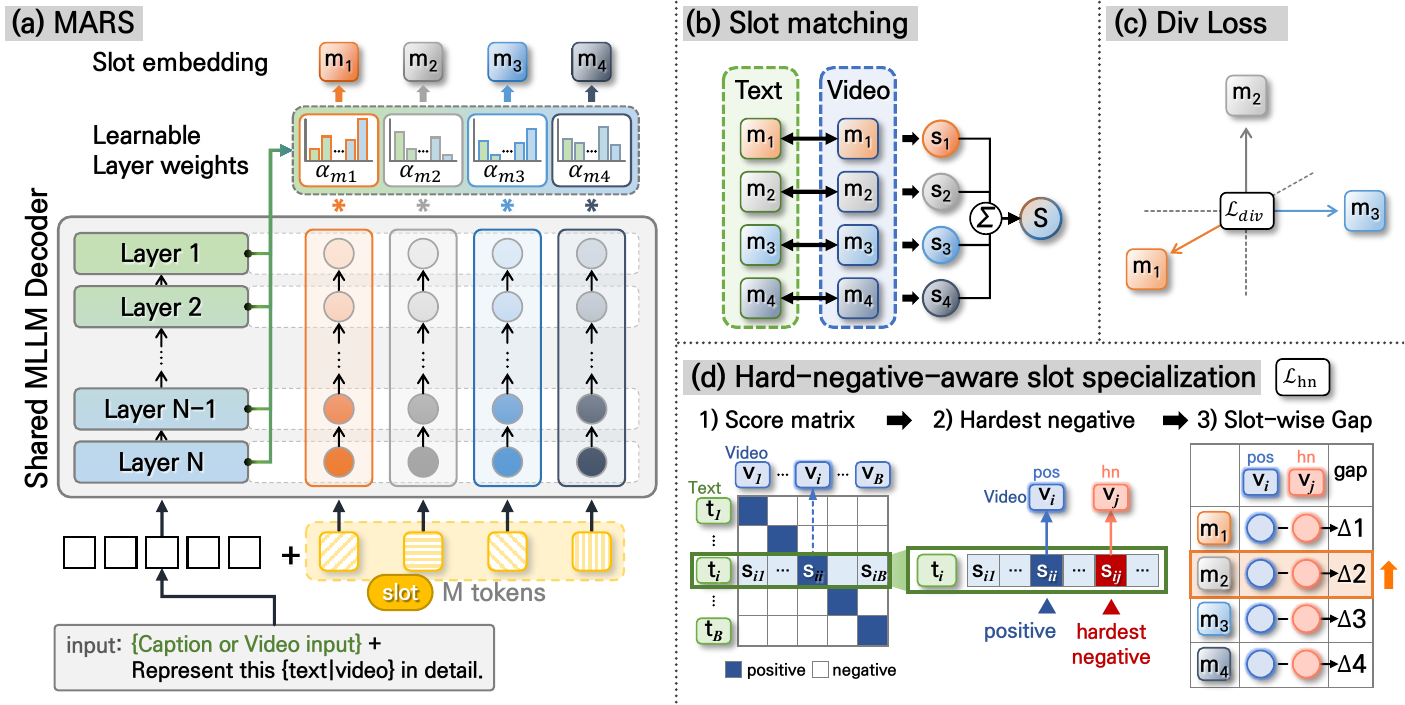}
    \caption{
    Overview of MARS.
    (a) The overall framework constructs multiple representation slots by appending adaptive representation tokens to the prompt and fusing their hidden states across decoder layers.
    (b) The resulting slots are matched between text and video to compute the retrieval score.
    (c) A diversity objective reduces redundancy among slots from the same input.
    (d) Hard-negative-aware slot specialization further strengthens discrimination against confusing negatives.
    }
    \label{fig:overview}
\end{figure*}

In this section, we introduce \textbf{MARS}, as illustrated in Figure~\ref{fig:overview}.
We first formulate slot-based text-video retrieval setting in Sec.~\ref{sec:problem_formulation}.
Next, Sec.~\ref{sec:mars} presents the main components of MARS, including representation token prompting, layer-fused slot construction, slot-wise matching, and hard-negative-aware slot specialization.
Finally, Sec.~\ref{sec:training_objective} defines the overall training objective.


\subsection{Problem Formulation}
\label{sec:problem_formulation}
Given a batch of paired texts and videos
\(\mathcal{B}=\{(t_i,v_i)\}_{i=1}^{B}\), text-video retrieval aims to assign
higher similarity scores to matched text-video pairs than to mismatched pairs.
We denote the similarity between text \(t_i\) and video \(v_j\) by \(S[i,j]\),
forming a score matrix \(S \in \mathbb{R}^{B \times B}\), where rows correspond
to texts and columns correspond to videos. 
For a unified notation, we let
\(a \in \{t,v\}\) denote the query side, with \(S^{t}=S\) for text-to-video retrieval and \(S^{v}=S^{\top}\) for video-to-text retrieval.

\subsection{MARS (Multi-layer Adaptive Representation Slots)}
\label{sec:mars}
By constructing multiple adaptive representation slots from hidden states across decoder layers, MARS preserves complementary fine-grained cues that may be weakened when each input is compressed into a single global vector.
\paragraph{Representation token prompting.}
For each input, we construct a chat-formatted prompt and place \(M\)
learnable adaptive representation tokens in the model response field, where
each token corresponds to one adaptive representation slot.
The prompt template is illustrated as follows:
\begin{tcolorbox}[
    colback=gray!5,
    colframe=black!70,
    boxrule=0.8pt,
    arc=3pt,
    left=4pt,
    right=4pt,
    top=3pt,
    bottom=3pt,
    width=\columnwidth,
    boxsep=2pt
]
\footnotesize
\setlength{\parskip}{0pt}
\textbf{system:} {\ttfamily You are a helpful assistant.}\\[2pt]
\textbf{user:} {\ttfamily \textcolor{green!45!black}{\{caption or video input\}}}\\
\hspace*{2.6em}{\ttfamily Represent this \{text|video\} in detail.}\\[2pt]
\textbf{assistant:} {\ttfamily \textcolor{green!45!black}{<slot\_1> <slot\_2> \(\cdots\) <slot\_M>}}
\end{tcolorbox}
The adaptive representation tokens are shared across text and video inputs. We denote them as \(\{\mathbf{e}^{\mathrm{slot}}_m\}_{m=1}^{M}\), where
\(\mathbf{e}^{\mathrm{slot}}_m \in \mathbb{R}^{D}\).

Since the MLLM follows causal attention, we use the hidden states immediately
preceding the adaptive representation tokens as slot representations.
Let \(\mathbf{h}_{i,m,t}^{(n)}, \mathbf{h}_{j,m,v}^{(n)} \in \mathbb{R}^{D}\)
denote the hidden states from the \(n\)-th decoder layer for the \(m\)-th slot
of text sample \(i\) and video sample \(j\), respectively.

\paragraph{Multi-layer adaptive slot construction.}
To construct each slot from hidden states across multiple decoder layers, we
apply slot-wise weighted layer fusion.
For the \(m\)-th slot, we use a learnable vector
\(\mathbf{w}_m \in \mathbb{R}^{N}\) to assign weights to different decoder layers.
The normalized layer weight \(\alpha_{m,n}\) is computed as:
\begin{equation}
\alpha_{m,n}
=
\frac{\exp(w_{m,n})}
{\sum_{n'=1}^{N} \exp(w_{m,n'})}.
\end{equation}
The hidden states from different layers are then fused to obtain the text and
video slot embeddings:
\begin{equation}
\mathbf{z}^{t}_{i,m}
=
\sum_{n=1}^{N}
\alpha_{m,n} \mathbf{h}_{i,m,t}^{(n)},
\quad
\mathbf{z}^{v}_{j,m}
=
\sum_{n=1}^{N}
\alpha_{m,n} \mathbf{h}_{j,m,v}^{(n)}.
\end{equation}
We apply \(L_2\)-normalization to each slot embedding, denoted by \(\hat{\mathbf{z}}\).

\paragraph{Slot-wise matching.}
After constructing the normalized text and video slot embeddings, we compare them in a slot-aligned manner.
For the \(m\)-th slot, we compute the slot-wise cosine similarity matrix:
\[
S_m[i,j]
=
\left\langle
\hat{\mathbf{z}}^{t}_{i,m},
\hat{\mathbf{z}}^{v}_{j,m}
\right\rangle .
\]
Applying the same notation to each slot-wise score matrix, we define \(S_m^{t}=S_m\) and
\(S_m^{v}=S_m^{\top}\). 
We then aggregate the slot-wise similarities across
\(M\) slots to obtain the final similarity score.
We use uniform aggregation, which is simple and performs consistently across benchmarks, as shown in Appendix~\ref{app:slot_aggregation}.
Accordingly, with slot weights
\(\beta_m = 1/M\), the aggregated score is defined as:
\begin{equation}
S[i,j]
=
\sum_{m=1}^{M}
\beta_m S_m[i,j].
\end{equation}

\paragraph{Hard-negative-aware slot specialization.}
Hard negatives are commonly used to strengthen cross-modal alignment by providing confusing mismatched samples~\citep{chen2020uniter,li2021albef}.
While prior methods mainly use them as pair-level training samples, we leverage them as slot-level signals for specialization.
We therefore introduce a hard-negative-aware slot specialization objective that encourages the most discriminative slot to separate the positive pair from the confusing negative.
For each query side \(a \in \{t,v\}\), we select the hardest negative using the corresponding score matrix:
\begin{equation}
j_a^*(i)
=
\arg\max_{j \neq i}
S^{a}[i,j].
\end{equation}
For each selected hard negative, we define the slot-wise positive-negative gap
as:
\begin{equation}
\Delta_m^{a}(i)
=
S_m^{a}[i,i] - S_m^{a}[i,j_a^*(i)].
\end{equation}
Here, the slot-wise gap measures how much the positive pair is separated from
the selected hard negative at slot \(m\).
We then take the largest slot-wise gap across slots:
\begin{equation}
\mathrm{gap}^{a}(i)
=
\max_{1 \leq m \leq M}
\Delta_m^{a}(i).
\end{equation}

This formulation selects the most discriminative slot for each hard negative; since the selected slot can vary across negatives, different slots are encouraged to specialize.
The query-side hinge loss is defined as:
\begin{equation}
\mathcal{L}_{\mathrm{hn}}^{a}
=
\frac{1}{B}
\sum_{i=1}^{B}
\max(0,\delta-\mathrm{gap}^{a}(i)),
\end{equation}
where \(\delta\) is the margin.
When the gap is smaller than \(\delta\), this term promotes a larger separation
between the positive pair and the selected hard negative.
The final hard-negative-aware slot specialization loss is obtained by averaging
over the text and video query sides:
\begin{equation}
\mathcal{L}_{\mathrm{hn}}
=
\frac{1}{2}
\sum_{a \in \{t,v\}}
\mathcal{L}_{\mathrm{hn}}^{a}.
\end{equation}

\subsection{Training Objective}
\label{sec:training_objective}

The model is trained with three objectives: symmetric contrastive alignment,
slot diversity regularization, and hard-negative-aware slot specialization.
For each side \(a \in \{t,v\}\), the contrastive loss is:
\begin{equation}
\mathcal{L}_{\mathrm{NCE}}^{a}
=
-\frac{1}{B}
\sum_{i=1}^{B}
\log
\frac{
\exp(\gamma S^{a}[i,i])
}{
\sum_{j=1}^{B}
\exp(\gamma S^{a}[i,j])
}.
\end{equation}
The symmetric InfoNCE loss is defined as:
\begin{equation}
\mathcal{L}_{\mathrm{NCE}}
=
\frac{1}{2}
\sum_{a \in \{t,v\}}
\mathcal{L}_{\mathrm{NCE}}^{a},
\end{equation}
where \(\gamma\) is the scaling factor.
Since multiple slots are extracted from the same input, different slots can
become redundant.
We therefore apply a squared cosine similarity regularizer within each side:
\begin{equation}
\mathcal{L}_{\mathrm{div}}^{a}
=
\frac{1}{B M(M-1)}
\sum_{i=1}^{B}
\sum_{m \neq m'}
\left[
(\hat{\mathbf{z}}^{a}_{i,m})^{\top}
\hat{\mathbf{z}}^{a}_{i,m'}
\right]^2.
\end{equation}
The total diversity loss is defined as:
\begin{equation}
\mathcal{L}_{\mathrm{div}}
=
\frac{1}{2}
\sum_{a \in \{t,v\}}
\mathcal{L}_{\mathrm{div}}^{a}.
\end{equation}
This term penalizes high similarity between slots from the same input and
encourages the slots to encode distinct retrieval-relevant information.

Together with the hard-negative-aware slot specialization loss defined above, the final training
objective is:
\begin{equation}
\mathcal{L}
=
\mathcal{L}_{\mathrm{NCE}}
+
\lambda_{\mathrm{div}}
\mathcal{L}_{\mathrm{div}}
+
\lambda_{\mathrm{hn}}
\mathcal{L}_{\mathrm{hn}},
\end{equation}
where \(\lambda_{\mathrm{div}}\) and \(\lambda_{\mathrm{hn}}\) are loss weights.

\definecolor{MarsBlue}{RGB}{224,236,255}
\definecolor{MarsSlate}{RGB}{232,238,250}
\definecolor{MarsTeal}{RGB}{219,242,235}
\definecolor{MarsMint}{RGB}{222,247,230}
\definecolor{MarsGreen}{RGB}{232,245,219}
\definecolor{MarsYellow}{RGB}{255,248,210}
\definecolor{MarsPeach}{RGB}{255,234,218}
\definecolor{MarsRose}{RGB}{255,229,234}
\definecolor{MarsLavender}{RGB}{240,230,255}
\definecolor{MarsGray}{RGB}{235,235,235}
\definecolor{MarsMintA}{RGB}{205,240,218}
\definecolor{MarsMintB}{RGB}{190,232,205}
\definecolor{MarsMintC}{RGB}{176,224,194}
\definecolor{MarsMintD}{RGB}{160,216,182}
\definecolor{MarsMintE}{RGB}{145,205,170}
\definecolor{MarsBlueA}{RGB}{205,224,255}
\definecolor{MarsBlueB}{RGB}{190,213,250}
\definecolor{MarsBlueC}{RGB}{176,202,245}
\definecolor{MarsBlueD}{RGB}{160,190,238}
\definecolor{MarsBlueE}{RGB}{145,178,230}
\definecolor{MarsDeepBlueA}{RGB}{190,215,255}
\definecolor{MarsDeepBlueB}{RGB}{170,200,250}
\definecolor{MarsDeepBlueC}{RGB}{150,185,245}
\definecolor{MarsDeepBlueD}{RGB}{130,170,235}
\definecolor{MarsDeepBlueE}{RGB}{110,155,225}
\definecolor{MarsCyanA}{RGB}{220,247,250}
\definecolor{MarsCyanB}{RGB}{200,239,244}
\definecolor{MarsCyanC}{RGB}{180,231,238}
\definecolor{MarsCyanD}{RGB}{160,224,232}
\definecolor{MarsCyanE}{RGB}{140,216,226}
\definecolor{MarsTabBlueA}{RGB}{225,240,250}
\definecolor{MarsTabBlueB}{RGB}{205,229,245}
\definecolor{MarsTabBlueC}{RGB}{185,218,240}
\definecolor{MarsTabBlueD}{RGB}{165,207,235}
\definecolor{MarsTabBlueE}{RGB}{145,196,230}

\definecolor{DirectBlock}{RGB}{235,240,248}
\definecolor{ScoringBlock}{RGB}{237,245,240}
\definecolor{MarsCyanLight}{RGB}{224,247,250}
\definecolor{MarsGreenLight}{RGB}{225,247,235}

\begin{table*}[t]
\centering
\scriptsize
\setlength{\tabcolsep}{1.45pt}
\renewcommand{\arraystretch}{1.08}
\caption{
Comparison with representative text-video retrieval methods on four benchmarks.
T2V and V2T denote text-to-video and video-to-text retrieval, respectively.
mR@1 denotes the mean R@1 across the four datasets.
$^{\dagger}$ denotes our reproduced InternVideo2-1B result.
$^{*}$ indicates the use of the DSL post-processing operation.
MARS-R denotes MARS with reranking.
}
\vspace{+3pt}
\label{tab:main_retrieval_results}
\begin{adjustbox}{max width=\textwidth}
\begin{tabular}{l|ccc|ccc|ccc|ccc|ccc|ccc|ccc|ccc|cc}
\toprule
\multirow{3}{*}{Method}
& \multicolumn{6}{c|}{DiDeMo}
& \multicolumn{6}{c|}{ActivityNet}
& \multicolumn{6}{c|}{LSMDC}
& \multicolumn{6}{c|}{MSR-VTT}
& \multicolumn{2}{c}{Avg.} \\
\cmidrule(lr){2-7}
\cmidrule(lr){8-13}
\cmidrule(lr){14-19}
\cmidrule(lr){20-25}
\cmidrule(lr){26-27}
& \multicolumn{3}{c|}{T2V}
& \multicolumn{3}{c|}{V2T}
& \multicolumn{3}{c|}{T2V}
& \multicolumn{3}{c|}{V2T}
& \multicolumn{3}{c|}{T2V}
& \multicolumn{3}{c|}{V2T}
& \multicolumn{3}{c|}{T2V}
& \multicolumn{3}{c|}{V2T}
& T2V & V2T \\
\cmidrule(lr){2-4}
\cmidrule(lr){5-7}
\cmidrule(lr){8-10}
\cmidrule(lr){11-13}
\cmidrule(lr){14-16}
\cmidrule(lr){17-19}
\cmidrule(lr){20-22}
\cmidrule(lr){23-25}
\cmidrule(lr){26-27}
& R@1 & R@5 & R@10
& R@1 & R@5 & R@10
& R@1 & R@5 & R@10
& R@1 & R@5 & R@10
& R@1 & R@5 & R@10
& R@1 & R@5 & R@10
& R@1 & R@5 & R@10
& R@1 & R@5 & R@10
& mR@1 & mR@1 \\
\midrule

\rowcolor{DirectBlock}
\multicolumn{27}{l}{\textbf{Direct similarity-based retrieval}} \\
\specialrule{0.45pt}{0pt}{0.25em}

CLIP4Clip~\cite{luo2022clip4clip}
& 42.8 & 68.5 & 79.2
& 42.5 & 70.6 & 80.2
& 40.5 & 72.4 & 83.4
& 42.6 & 73.4 & 85.6
& 21.6 & 41.8 & 49.8
& 20.9 & 40.7 & 49.1
& 44.5 & 71.4 & 81.6
& 43.1 & 70.5 & 81.2
& 37.4 & 37.3 \\

CLIP-ViP~\cite{xue2023clipvip}
& 48.6 & 77.1 & 84.4
& -- & -- & --
& 51.1 & 78.4 & 88.3
& -- & -- & --
& 25.6 & 45.3 & 54.4
& -- & -- & --
& 50.1 & 74.8 & 84.6
& -- & -- & --
& 43.9 & -- \\

ViCLIP~\cite{viclip}
& 49.4 & -- & --
& 50.2 & -- & --
& 49.8 & -- & --
& 48.1 & -- & --
& 33.0 & -- & --
& 32.5 & -- & --
& 52.5 & -- & --
& 51.8 & -- & --
& 46.2 & 45.6 \\

Video-ColBERT~\cite{reddy2025videocolbert}
& 48.2 & 75.1 & 83.7
& -- & -- & --
& 45.5 & 74.6 & 85.5
& -- & -- & --
& -- & -- & --
& -- & -- & --
& 48.1 & 74.9 & 83.9
& -- & -- & --
& -- & -- \\

InternVideo~\cite{wang2022internvideo}
& 57.9 & 82.4 & 88.9
& 59.1 & 81.8 & 89.0
& 62.2 & 85.9 & 93.2
& 62.8 & 86.2 & 93.3
& 34.0 & 53.7 & 62.9
& 34.9 & 54.6 & 63.1
& 55.2 & 79.6 & 87.5
& 57.9 & 79.2 & 86.4
& 52.3 & 53.7 \\

\addlinespace[0.15em]

\rowcolor{MarsCyanLight}
\textbf{\makebox[3.4em][l]{MARS} (Ours)}
& \underline{79.7} & \underline{92.3} & \underline{95.5}
& \underline{75.7} & \underline{92.4} & \underline{95.4}
& \underline{75.7} & \underline{93.0} & \underline{96.8}
& \underline{73.2} & \underline{92.6} & \underline{97.0}
& \underline{51.0} & \underline{70.9} & \underline{78.8}
& \underline{50.0} & \underline{71.5} & \underline{79.0}
& \underline{61.6} & \underline{83.6} & \underline{89.1}
& \underline{59.0} & \underline{85.0} & \underline{91.0}
& \underline{67.0} & \underline{64.5} \\

\rowcolor{MarsCyanLight}
\textbf{\makebox[3.4em][l]{MARS$^{*}$} (Ours)}
& \textbf{84.6} & \textbf{94.6} & \textbf{97.0}
& \textbf{84.2} & \textbf{94.4} & \textbf{97.1}
& \textbf{82.1} & \textbf{95.1} & \textbf{98.0}
& \textbf{82.1} & \textbf{95.6} & \textbf{98.1}
& \textbf{54.4} & \textbf{73.9} & \textbf{82.0}
& \textbf{54.9} & \textbf{74.4} & \textbf{81.4}
& \textbf{65.5} & \textbf{85.8} & \textbf{91.1}
& \textbf{66.8} & \textbf{86.7} & \textbf{92.7}
& \textbf{71.6} & \textbf{72.0} \\

\specialrule{0.9pt}{0.6em}{0.25em}

\rowcolor{ScoringBlock}
\multicolumn{27}{l}{\textbf{Cross-modal matching and reranking}} \\
\specialrule{0.45pt}{0pt}{0.25em}

UMT~\cite{li2023umt}
& 70.4 & 90.1 & 93.5
& 67.9 & 88.6 & 93.0
& 66.8 & 89.1 & 94.9
& 64.4 & 89.1 & 94.8
& 43.0 & 65.5 & 73.0
& 41.4 & 64.3 & 71.5
& 58.8 & 81.0 & 87.1
& 58.6 & 81.6 & 86.5
& 59.8 & 58.1 \\

InternVideo2 1B$^{\dagger}$$^{*}$~\cite{wang2024internvideo2}
& 72.8 & 90.3 & 93.9
& 69.1 & 89.1 & 93.7
& 66.2 & 88.5 & 94.3
& 62.8 & 86.7 & 93.5
& 42.7 & 64.7 & 72.9
& 42.7 & 63.7 & 71.3
& 58.7 & 80.2 & 87.0
& 56.8 & 79.3 & 85.6
& 60.1 & 57.8 \\

InternVideo2 6B$^{*}$~\cite{wang2024internvideo2}
& 74.2 & -- & --
& 71.9 & -- & --
& 74.1 & -- & --
& 68.7 & -- & --
& 46.4 & -- & --
& 46.7 & -- & --
& 62.8 & -- & --
& 60.2 & -- & --
& 64.4 & 61.9 \\

BLiM~\cite{ko2025blim}
& \underline{86.4} & \underline{95.6} & \underline{96.4}
& \textbf{82.8} & \textbf{95.6} & \underline{96.4}
& \underline{81.0} & \underline{94.2} & \underline{96.6}
& \underline{74.4} & \underline{92.6} & \underline{96.2}
& \underline{55.7} & \underline{73.1} & \underline{78.2}
& \underline{49.1} & \underline{71.0} & \underline{77.1}
& \underline{64.7} & \underline{83.9} & \underline{88.2}
& \textbf{62.2} & \underline{82.7} & \underline{87.0}
& \underline{72.0} & \underline{67.1} \\

\addlinespace[0.15em]

\rowcolor{MarsGreenLight}
\textbf{MARS-R (Ours)}
& \textbf{87.6} & \textbf{96.1} & \textbf{96.8}
& \underline{82.2} & \underline{94.9} & \textbf{96.5}
& \textbf{83.5} & \textbf{95.4} & \textbf{97.9}
& \textbf{77.9} & \textbf{94.3} & \textbf{97.6}
& \textbf{56.1} & \textbf{74.1} & \textbf{80.9}
& \textbf{50.0} & \textbf{71.5} & \textbf{79.0}
& \textbf{65.8} & \textbf{84.8} & \textbf{90.0}
& \underline{61.7} & \textbf{85.1} & \textbf{91.1}
& \textbf{73.2} & \textbf{67.9} \\

\bottomrule
\end{tabular}
\end{adjustbox}
\vspace{+3pt}
\end{table*}

\begin{table*}[t]
\centering
\scriptsize
\setlength{\tabcolsep}{2.6pt}
\renewcommand{\arraystretch}{1.05}

\providecommand{\yes}{\checkmark}
\providecommand{\no}{\textcolor{gray!55}{--}}
\providecommand{\Ldiv}{\mathcal{L}_{\mathrm{div}}}
\providecommand{\Lhn}{\mathcal{L}_{\mathrm{hn}}}

\begin{minipage}[t]{0.50\textwidth}
\centering
\caption{
Component ablation. LF and Slots denote multi-layer fusion and adaptive representation slots.
}
\label{tab:ablation_components_all}

\begin{adjustbox}{max width=\linewidth}
\begin{tabular}{cccc|cc|cc|cc|cc}
\toprule
\multicolumn{4}{c|}{Components}
& \multicolumn{2}{c|}{DiDeMo}
& \multicolumn{2}{c|}{ActivityNet}
& \multicolumn{2}{c|}{LSMDC}
& \multicolumn{2}{c}{MSR-VTT} \\
\cmidrule(lr){1-4}
\cmidrule(lr){5-6}
\cmidrule(lr){7-8}
\cmidrule(lr){9-10}
\cmidrule(lr){11-12}
LF & Slots & $\Ldiv$ & $\Lhn$
& T2V & V2T
& T2V & V2T
& T2V & V2T
& T2V & V2T \\
\midrule

\no & \no & \no & \no
& 72.9 & 71.5
& 69.5 & 66.0
& 47.5 & 47.1
& 58.3 & 56.7 \\

\yes & \no & \no & \no
& 76.6 & 74.9
& 72.7 & 70.0
& 47.9 & 46.6
& 59.8 & 57.9 \\

\yes & \yes & \no & \no
& 77.4 & 75.0
& 75.6 & \textbf{73.6}
& 49.9 & 50.2
& 60.8 & 57.9 \\

\yes & \yes & \yes & \no
& 77.5 & 75.0
& \textbf{75.7} & 73.2
& 50.9 & 49.7
& 61.1 & \textbf{59.0} \\

\rowcolor{MarsCyanA}
\yes & \yes & \yes & \yes
& \textbf{79.7} & \textbf{75.7}
& \textbf{75.7} & 73.2
& \textbf{51.0} & 50.0
& \textbf{61.6} & \textbf{59.0} \\

\bottomrule
\end{tabular}
\end{adjustbox}
\end{minipage}
\hfill
\begin{minipage}[t]{0.46\textwidth}
\centering
\small
\setlength{\tabcolsep}{3.0pt}
\renewcommand{\arraystretch}{0.9}
\vspace{0pt}
\caption{
Effect of the number of adaptive representation slots.
}
\label{tab:ablation_num_slots}

\begin{tabularx}{\linewidth}{
>{\centering\arraybackslash}X
>{\centering\arraybackslash}X
>{\centering\arraybackslash}X
>{\centering\arraybackslash}X}
\toprule
\textbf{Slots \(M\)} & \textbf{T2V} & \textbf{V2T} & \textbf{Avg.} \\
\midrule
2 & 76.5 & 74.8 & 75.6 \\
3 & 78.8 & 73.9 & 76.3 \\
\rowcolor{MarsCyanA}
4 & 79.7 & \textbf{75.7} & \textbf{77.7} \\
5 & \textbf{79.9} & 74.9 & 77.4 \\
6 & 77.3 & 74.6 & 75.9 \\
\bottomrule
\end{tabularx}
\end{minipage}

\end{table*}

\section{Experiments}
\label{sec:experiments}

\subsection{Experimental Setup}
\label{sec:exp_setup}

\paragraph{Datasets and metrics.}
We use DiDeMo, ActivityNet, LSMDC, and MSR-VTT~\citep{hendricks2017localizing,krishna2017dense,rohrbach2017movie,xu2016msrvtt}, covering diverse domains, video lengths, and caption styles.
For each benchmark, we report Recall@K (R@1, R@5, and R@10) for text-to-video (T2V) and video-to-text (V2T) retrieval, along with mR@1 across the four benchmarks.
Dataset statistics and preprocessing details are provided in Appendix~\ref{app:dataset_details}.

\paragraph{Implementation details.}
MARS uses VideoChat-Flash-Qwen2-7B~\citep{videochat_flash} as the MLLM, containing a UMT-L vision encoder~\citep{li2023umt}, a linear projection, and a 28-layer Qwen2~\citep{qwen2}. Pretrained weights remain frozen during training. We only update adaptive representation tokens, slot-wise layer-fusion weights, and LoRA~\citep{lora} parameters (applied to the projection and Qwen2). Unless specified, we set $M=4$ for DiDeMo, ActivityNet, and LSMDC, and $M=3$ for MSR-VTT, with $\lambda_{\mathrm{div}}=0.1$, $\lambda_{\mathrm{hn}}=0.05$, and $\delta=0.04$. Additional details are provided in Appendix~\ref{sec:supp_implementation}.

\begin{figure*}[t]
\centering
\includegraphics[width=\textwidth]{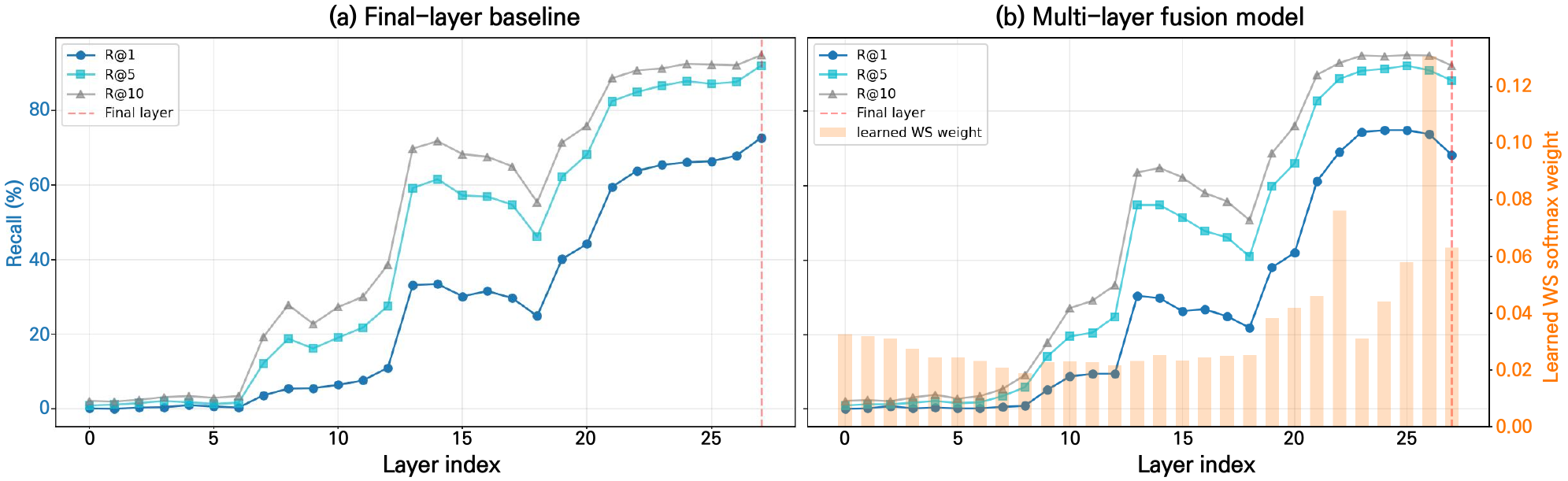}
\caption{
Layer-wise retrieval performance.
(a) evaluates the final-layer baseline across decoder layers.
(b) reports the multi-layer fusion model, where bars show the learned layer-fusion weights.
The dashed line marks the final layer.
}
\label{fig:layerwise_recall}
\end{figure*}

\begin{figure*}[t]
\centering
\includegraphics[width=0.91\textwidth]{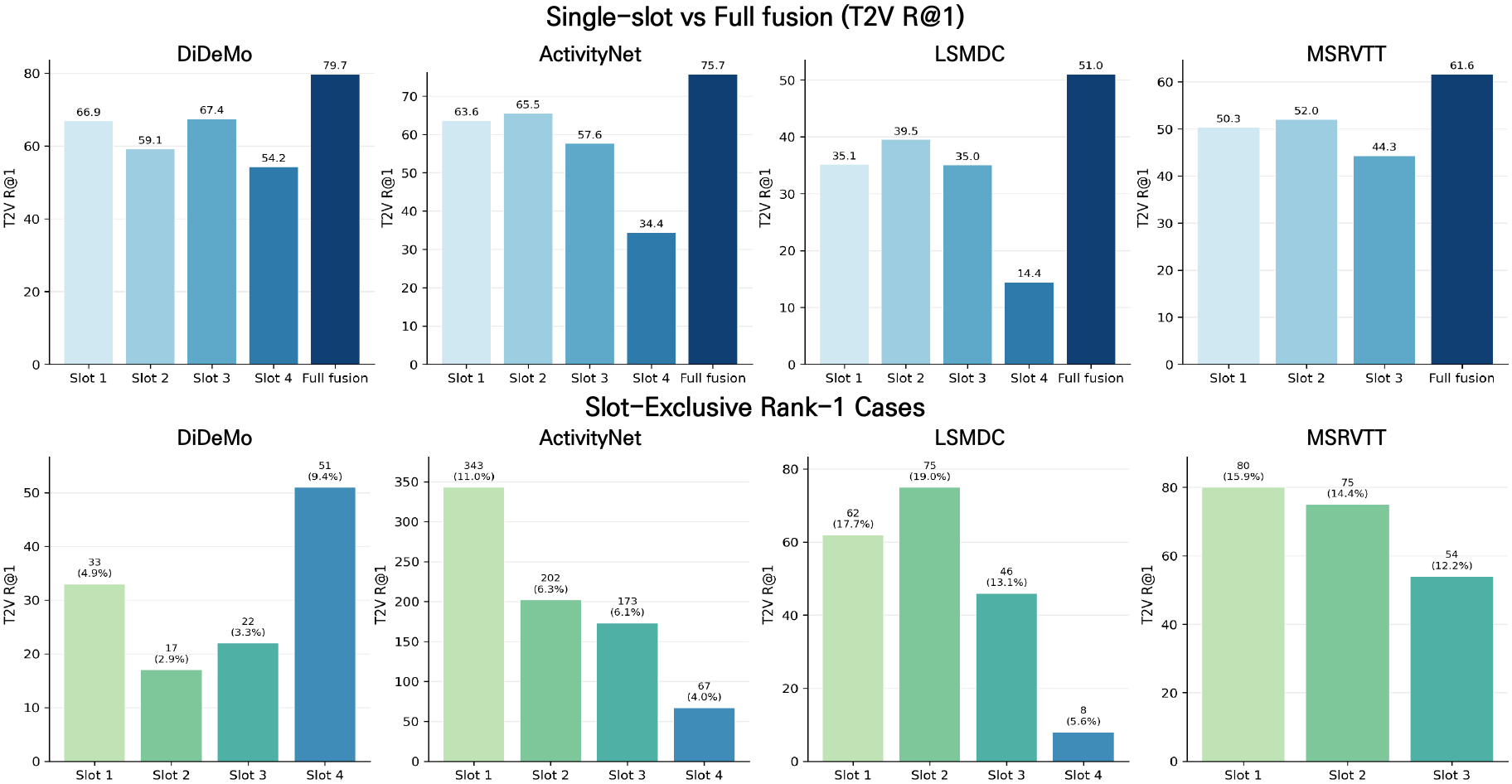}
\caption{
Slot-level retrieval analysis across benchmarks.
Uniform aggregation outperforms each individual slot, and slot-exclusive rank-1 cases indicate complementary retrieval behavior among slots.
}
\label{fig:slot_analysis_retrieval}
\end{figure*}





\subsection{Main Results}
\label{sec:main_results}

The compared methods fall into two retrieval protocols: direct similarity-based retrieval and cross-modal matching  or reranking.
Direct similarity methods encode videos and texts independently, and rank candidates by representation similarity.
In contrast, cross-modal matching and reranking methods  retrieve candidates via embedding similarity and refine scores through additional cross-modal interaction or candidate-level reranking. Specifically, MARS-R first retrieves candidates using MARS and then reranks them with BLiM.

As shown in Table~\ref{tab:main_retrieval_results}, MARS achieves the strongest performance among direct similarity-based retrieval methods across four benchmarks.
Without cross-modal reranking, MARS outperforms InternVideo2-6B$^{*}$, which relies on cross-modal matching, while retaining a direct similarity-based retrieval structure, achieving 67.0 T2V mR@1 and 64.5 V2T mR@1.
This suggests that multi-layer evidence aggregation and complementary slot-wise matching produce stronger retrieval embeddings for direct matching.
Furthermore, with the same DSL~\cite{cheng2021improving} post-processing, which calibrates retrieval scores using dual softmax normalization, MARS$^{*}$ improves to 71.6 T2V mR@1 and 72.0 V2T mR@1, indicating that the score distribution produced by MARS also benefits from score calibration. When candidate-level reranking is applied, MARS-R achieves the best T2V mR@1 of 73.2 and improves over BLiM in both average T2V and V2T performance.

\begin{figure*}[!t]
\centering
\includegraphics[width=0.91\textwidth]{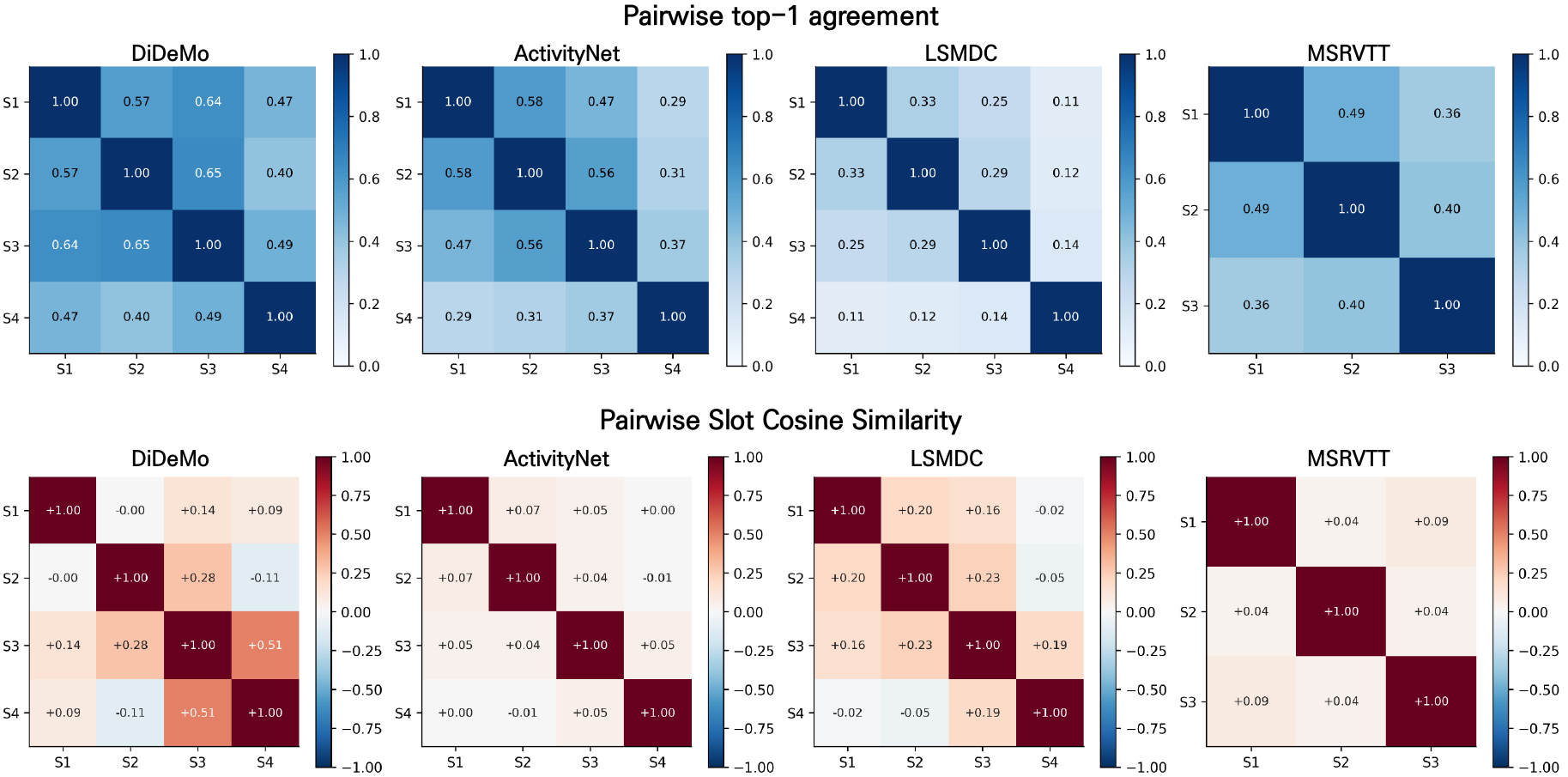}

\caption{
Diversity analysis of adaptive representation slots.
Pairwise top-1 agreement and slot embedding cosine similarity show that the learned slots make different retrieval decisions and remain distinct in the embedding space.
}
\label{fig:slot_analysis_diversity}

\end{figure*}

\subsection{Ablation Studies}
\label{sec:ablation}

We conduct ablations across four benchmarks to isolate each component of MARS.
Tables~\ref{tab:ablation_components_all} and~\ref{tab:ablation_num_slots} report R@1 for T2V and V2T retrieval.
Table~\ref{tab:ablation_components_all} starts from a final-layer single-slot baseline and progressively adds layer fusion, representation slots, slot diversity regularization, and hard-negative-aware slot specialization.
\textbf{Multi-layer fusion:} This yields the largest individual gain, increasing the average R@1 across the eight T2V/V2T settings from 61.2 to 63.3.
This supports our motivation that retrieval cues are distributed across decoder depths rather than concentrated in the final layer.
\textbf{Adaptive representation slots:} Adding multiple slots further raises the average R@1 to 65.1, showing that they preserve complementary matching information that may be overly compressed in a single representation.
\textbf{Slot diversity regularization:} This provides a modest additional gain, suggesting that reducing redundancy among slots helps maintain distinct retrieval cues.
\textbf{Hard-negative-aware slot specialization:} This final component improves the full model to the best average R@1 of 65.7, with particularly strong gains on DiDeMo and consistent improvements in T2V retrieval across all four benchmarks.
Together, these results show that the proposed components provide complementary gains over the final-layer single-slot baseline.

We further vary the number of slots \(M\) in Table~\ref{tab:ablation_num_slots}.
The average R@1 peaks at 77.7 with $M=4$ (up from 75.6 at $M=2$), but drops with more slots.
This indicates that a moderate number of slots is optimal, whereas excessive slots may introduce redundant or less discriminative representations.

\subsection{Dissecting Representation Slots}
\label{sec:analysis}

Beyond performance, we analyze why representation slots improve text-video retrieval.
We probe three properties that underlie this improvement:
1) \textbf{layer-wise evidence}: whether retrieval-relevant cues are distributed across decoder layers rather than concentrated in the final layer;
2) \textbf{slot complementarity}: whether different slots capture complementary matching cues; and
3) \textbf{slot distinctness}: whether the learned slots remain distinct in retrieval behavior and embedding space.

\paragraph{Q1. Where is retrieval evidence encoded across decoder layers?}
Figure~\ref{fig:layerwise_recall} compares single-layer retrieval performance across decoder layers between the final-layer baseline and the multi-layer fusion model.
In the final-layer baseline, R@1 increases toward the final decoder layer, where the contrastive objective is directly applied.
However, this pattern changes when multi-layer fusion is trained.
The strongest single-layer performance is no longer limited to the final layer; instead, several upper decoder layers achieve comparable R@1.

Interestingly, the learned fusion weights do not simply follow the single-layer R@1 curve.
Some layers with lower individual R@1 still receive non-negligible weights, suggesting that they provide information that is useful when combined with other layers.
This indicates that retrieval evidence is distributed across decoder layers, supporting our design choice of constructing slot embeddings from layer-fused representations rather than relying only on the final decoder layer.

\begin{table*}[!t]
\centering
\scriptsize
\setlength{\tabcolsep}{5.0pt}
\renewcommand{\arraystretch}{1.05}

\caption{
Computational cost comparison across four benchmarks.
Offline cost is measured per video, while online cost is measured per query for each benchmark.
Total computation is reported in million GFLOPs (M GFLOPs), and inference time is measured on a single NVIDIA A100 GPU.
}
\label{tab:inference_cost}

\resizebox{\textwidth}{!}{
\begin{tabular}{l c cccc c c c}
\toprule

\multirow{2}{*}{\textbf{Method}}
& \multirow{2}{*}{\makecell{\textbf{Offline}\\\textbf{GF/video}}}
& \multicolumn{4}{c}{\textbf{Online GF/query}}
& \multirow{2}{*}{\makecell{\textbf{Total}\\\textbf{(M GFLOPs)}}}
& \multirow{2}{*}{\textbf{Time}}
& \multirow{2}{*}{\textbf{mR@1}} \\

\cmidrule(lr){3-6}

&
& \textbf{DiDeMo}
& \textbf{ActivityNet}
& \textbf{LSMDC}
& \textbf{MSR-VTT}
&
&
& \\

\midrule

UMT
& 267.8
& 1,740.8
& 2,683.4
& 2,089.2
& 1,395.2
& 20.5
& 2,351.8s
& 59.8 \\

InternVideo2-1B
& 2,542.6
& 4,576.1
& 5,661.3
& 4,977.5
& 4,177.4
& 61.7
& 8,717.4s
& 60.1 \\

InternVideo2-6B
& 13,384.1
& 2,857.2
& 3,942.5
& 3,258.7
& 2,458.5
& 134.0
& 10,424.3s
& 64.4 \\

\rowcolor{MarsCyanA}
\textbf{MARS (Ours)}
& \textbf{5,657.7}
& \textbf{322.3}
& \textbf{1,031.8}
& \textbf{659.3}
& \textbf{274.3}
& \textbf{51.1}
& \textbf{3,626.1s}
& \textbf{67.0} \\

\bottomrule
\end{tabular}
}
\vspace{-0.5em} 
\end{table*}

\begin{table*}[!t]
\centering
\scriptsize
\setlength{\tabcolsep}{5.0pt}
\renewcommand{\arraystretch}{1.05}

\caption{
Generalization of MARS across MLLM backbones and model scales.
Each entry reports T2V / V2T R@1.
$\dagger$ denotes the backbone equipped with MARS.
$\Delta$ denotes the average absolute R@1 improvement over the corresponding base model
across both retrieval directions and all four benchmarks.
}
\label{tab:mllm_backbone}

\vspace{-0.5em} 

\resizebox{\textwidth}{!}{
\begin{tabular}{lcccccc}
\toprule

\textbf{Model}
& \textbf{DiDeMo}
& \textbf{ActivityNet}
& \textbf{LSMDC}
& \textbf{MSR-VTT}
& \textbf{Avg.}
& \textbf{$\Delta$} \\

\midrule

VideoLLaMA3-7B
& 61.0 / 56.8
& 60.4 / 58.6
& 38.8 / 36.9
& 54.1 / 53.5
& 53.6 / 51.5
& -- \\

\rowcolor{MarsCyanA}
\textbf{VideoLLaMA3-7B$^\dagger$ (Ours)}
& \textbf{63.8 / 61.1}
& \textbf{66.4 / 64.7}
& \textbf{39.6 / 38.4}
& \textbf{56.1 / 54.3}
& \textbf{56.5 / 54.6}
& \textbf{+3.0} \\

\midrule

Qwen2-VL-7B
& 57.7 / 56.8
& 53.3 / 52.4
& 34.3 / 32.0
& 54.8 / 52.8
& 50.0 / 48.5
& -- \\

\rowcolor{MarsCyanA}
\textbf{Qwen2-VL-7B$^\dagger$ (Ours)}
& \textbf{62.1 / 60.8}
& \textbf{58.9 / 57.7}
& \textbf{36.1 / 32.3}
& \textbf{56.3 / 54.7}
& \textbf{53.4 / 51.4}
& \textbf{+3.1} \\

\midrule

Qwen2-VL-2B
& 54.3 / 52.9
& 48.5 / 46.6
& 30.5 / 28.7
& 52.3 / 48.1
& 46.4 / 44.1
& -- \\

\rowcolor{MarsCyanA}
\textbf{Qwen2-VL-2B$^\dagger$ (Ours)}
& \textbf{57.1 / 56.2}
& \textbf{51.9 / 50.5}
& \textbf{31.4 / 28.9}
& \textbf{53.3 / 50.6}
& \textbf{48.4 / 46.6}
& \textbf{+2.3} \\

\bottomrule
\end{tabular}
}
\vspace{-0.8em} 
\end{table*}

\paragraph{Q2. Do multiple slots provide complementary matching cues?}
We next examine whether the learned slots capture complementary retrieval evidence.
Figure~\ref{fig:slot_analysis_retrieval} compares the single-slot T2V performance of each slot with uniform aggregation.
Across all benchmarks, aggregation significantly outperforms the strongest individual slot (R@1 gains: +12.3 on DiDeMo, +10.2 ActivityNet, +11.5 LSMDC, +9.6 MSR-VTT), indicating the final score benefits from combined signals rather than a single dominant slot.
This indicates that the final retrieval score is not dominated by a single slot, but benefits from combining multiple slot-wise matching signals.

To examine whether this aggregation gain reflects complementary successes across slots, we count slot-exclusive rank-1 cases, where only one slot retrieves the correct video at rank one.
Such cases appear consistently across benchmarks (123 on DiDeMo, 785 ActivityNet, 191 LSMDC, 209 MSR-VTT).
This shows that different slots can recover different correct matches rather than simply duplicating the strongest slot.
Therefore, a slot should not be judged only by its standalone accuracy; even a weaker slot can contribute by resolving retrieval cases that other slots miss.

\paragraph{Q3. Do learned slots form distinct representations?}
Figure~\ref{fig:slot_analysis_diversity} examines whether the learned slots collapse into similar representations.
The mean off-diagonal agreement remains well below full agreement across benchmarks (0.537 on DiDeMo, 0.429 ActivityNet, 0.206 LSMDC, 0.419 MSR-VTT), showing that different slots often produce different top-ranked candidates rather than repeatedly retrieving the same video.

The bottom row reports pairwise cosine similarity between video-side slot embeddings.
The off-diagonal similarities remain generally small, indicating that the slots are also separated in the embedding space.
Together, the retrieval-level agreement and embedding-space analyses show that the learned slots do not collapse into redundant copies, but maintain distinct representations that support multi-slot retrieval.

\subsection{Computational Cost Analysis}
\label{sec:inference_cost}
Since MARS builds on an MLLM, inference cost is an important practical consideration.
We analyze the computational cost across four benchmarks by separating offline video encoding from online retrieval-time computation, as reported in Table~\ref{tab:inference_cost}.
The offline cost comes from video encoding, which can be precomputed and stored, whereas the online cost is incurred per text query for text encoding and similarity computation.

After video representations are precomputed, MARS follows a dual-encoder retrieval structure: it encodes each text query once and computes similarity scores against stored video representations.
By contrast, UMT and InternVideo2 include cross-modal matching components in their retrieval pipelines, resulting in substantially higher online computation.
MARS achieves the lowest online cost across all four benchmarks, while obtaining the highest average T2V R@1 of 67.0.
Compared with InternVideo2, MARS also requires less total computation and evaluation time despite achieving stronger retrieval performance.
These results show that MARS provides an effective balance between retrieval accuracy and online computational efficiency.

\subsection{Generalization Across MLLM Backbones}
\label{sec:mllm_backbone}

To examine whether the effectiveness of MARS generalizes across different architectures, we evaluate it on three distinct MLLM backbones: VideoLLaMA3-7B~\cite{zhang2025videollama3} and Qwen2-VL~\cite{wang2024qwen2vl} (7B and 2B). For each backbone, we compare MARS against its corresponding single-representation baseline. As shown in Table~\ref{tab:mllm_backbone}, MARS consistently improves both T2V and V2T R@1 across all four benchmarks. Notably, it yields consistent absolute improvements in average R@1 ranging from 2.0\% to 3.4\% across VideoLLaMA3-7B, Qwen2-VL-7B, and Qwen2-VL-2B for both retrieval directions. These robust enhancements confirm that MARS is not tied to a specific backbone, but serves as a general mechanism for extracting stronger retrieval representations from diverse MLLMs.

\begin{table}[t]
\centering
\vspace{-0.4em}
\scriptsize
\setlength{\tabcolsep}{2.5pt}
\renewcommand{\arraystretch}{1.08}

\setlength{\MarsTableWidth}{1.03\columnwidth}

\caption{
Text-image retrieval on COCO and Flickr30K. All values are R@1.
}
\label{tab:text_image_retrieval}
\vspace{-0.4em}

\makebox[\columnwidth][c]{%
\begin{tabularx}{\MarsTableWidth}{
l
>{\centering\arraybackslash}X
>{\centering\arraybackslash}X
>{\centering\arraybackslash}X
>{\centering\arraybackslash}X
}

\toprule

\multirow{2}{*}{\textbf{Method}}
& \multicolumn{2}{c}{\textbf{COCO (FT)}}
& \multicolumn{2}{c}{\textbf{Flickr30K (ZS)}} \\

\cmidrule(lr){2-3}
\cmidrule(lr){4-5}

& \textbf{T2I}
& \textbf{I2T}
& \textbf{T2I}
& \textbf{I2T} \\

\midrule

\multicolumn{5}{l}{\textit{Direct retrieval}} \\[-0.2em]

BEiT-3~\cite{wang2023beit3}
& 65.1
& 82.7
& \textbf{89.1}
& \textbf{97.5} \\

\FullRowColor{MarsCyanA}%
\textbf{MARS (Ours)}
& \textbf{66.3}
& \textbf{83.9}
& \textbf{89.1}
& 97.0 \\

\midrule

\multicolumn{5}{l}{\textit{Fusion reranking}} \\[-0.2em]

ALBEF~\cite{li2021albef}
& 60.7
& 77.6
& 82.8
& 94.1 \\

BLIP~\cite{li2022blip}
& 65.1
& 82.4
& 86.7
& 96.7 \\

BLIP-2~\cite{li2023blip2}
& \textbf{68.3}
& \textbf{85.4}
& \textbf{89.7}
& \textbf{97.6} \\
\bottomrule
\end{tabularx}%
}
\end{table}

\subsection{Generalization to Text--Image Retrieval}
\label{sec:text_image_generalization}

Although our main experiments focus on text-video retrieval, the MARS framework is not restricted to video inputs. To examine its applicability beyond the video domain, we evaluate MARS on text-image retrieval using COCO~\cite{lin2014coco} (fine-tuned) and Flickr30K~\cite{plummer2015flickr30k} (zero-shot), following the Karpathy split~\cite{karpathy2015deep} adopted by BEiT-3. As shown in Table~\ref{tab:text_image_retrieval}, MARS achieves R@1 scores 1.2 points higher than BEiT-3 on COCO in both T2I and I2T directions. On Flickr30K, MARS maintains highly competitive zero-shot performance. Although BLIP-2 achieves higher scores with an additional cross-modal reranking stage, MARS remains competitive using only direct similarity-based retrieval. These results indicate that the core MARS mechanism generalizes well beyond the video domain, serving as a powerful representation extractor for text-image retrieval as well.

\subsection{Qualitative Results}
\label{sec:qualitative_results}

\begin{figure}[t]
\centering
\includegraphics[width=\columnwidth]{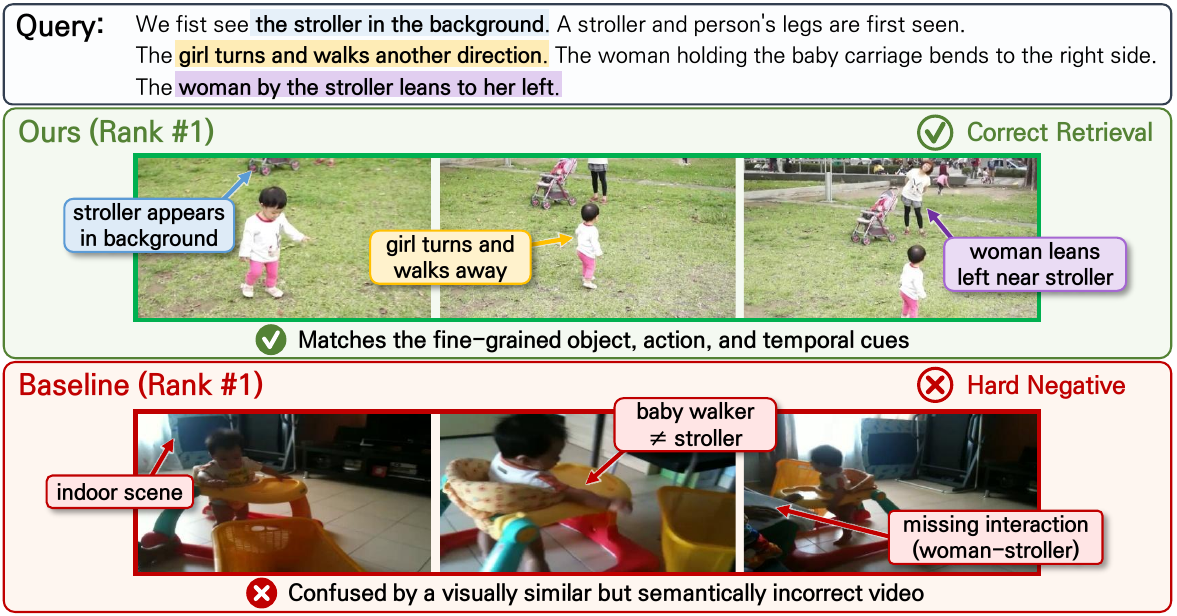}
\caption{
Qualitative hard-negative comparison. Unlike the baseline, MARS correctly ranks the ground truth above a visually similar negative.
}
\vspace{-0.8em}
\label{fig:qualitative_result}
\end{figure}

Figure~\ref{fig:qualitative_result} shows a qualitative hard-negative case from DiDeMo.
For the same text query, the final-layer single-slot baseline ranks a visually plausible negative above the ground-truth video.
We use this example to examine whether MARS can correct confusion between candidates with similar scene-level content.
The query contains several fine-grained cues, including the stroller, the girl's movement, and the directional motions of nearby women.
While the baseline assigns a higher score to the hard negative, MARS reverses this ordering and retrieves the ground-truth video at rank one.
This example illustrates that MARS can better distinguish the ground-truth video from a confusing negative using the final score.

\section{Conclusion}
\label{sec:conclusion}

In this paper, we introduced MARS, a multi-layer and multi-slot embedding framework for text-video retrieval with MLLMs. Through this design, experiments on four benchmarks show that MARS achieves strong retrieval performance while retaining a direct similarity-based retrieval structure. Further ablations and analyses confirm that multi-layer fusion, adaptive representation slots, and hard-negative-aware specialization provide complementary gains. These findings suggest that richer representation extraction from MLLMs is promising for fine-grained retrieval. Future work can explore adaptive slot interactions, temporal reasoning, and broader retrieval scenarios.

\section*{Limitations}
\label{sec:limitations}

This work has several limitations. First, although MARS is evaluated across multiple established text-video retrieval benchmarks, these datasets may not fully reflect open-domain scenarios involving substantially longer videos, noisy descriptions, or diverse user queries. Second, while we further demonstrate the generalization of MARS from text-video to text-image retrieval, its applicability to other modalities, such as audio-language or 3D-language retrieval, remains unexplored. Third, our study focuses on video-level retrieval and does not explicitly address temporal grounding or moment-level retrieval. Extending MARS to these broader retrieval settings represents a promising direction for future work.

\paragraph{Artifact licenses.}
We use publicly available benchmark datasets and pretrained models for research purposes, following their respective licenses and terms of use.

\section*{Acknowledgements}
This work was partly supported by Institute of Information and communications Technology Planning \& Evaluation (IITP) under the Development of Multimodal Data Input-Based Search Augmentation Generation Technology (IITP-2026-RS-2024-00455244, 50\%), the Leading Generative AI Human Resources Development (IITP-2026-RS-2026-25544647, 25\%) and the Artificial Intelligence Innovation Human Resources Development (IITP-2026-RS-2026-25549817, 25\%), grant funded by the Korea government(MSIT).

\bibliography{custom}

\clearpage
\appendix

\setcounter{table}{0}
\renewcommand{\thetable}{A\arabic{table}}

\setcounter{figure}{0}
\renewcommand{\thefigure}{A\arabic{figure}}

\section*{Appendix Overview}

\begin{itemize}
\item In Appendix~\ref{app:dataset_details}, we provide dataset details for the text-video retrieval benchmarks used in our experiments.
\item In Appendix~\ref{sec:supp_implementation}, we describe the implementation details, including model architecture, trainable parameters, and training configuration.
\item In Appendix~\ref{app:slot_aggregation}, we analyze slot aggregation strategies and justify the use of uniform aggregation.
\item In Appendix~\ref{app:slot_analysis}, we present additional analyses of adaptive representation slots using slot geometry diagnostics.
\end{itemize}

\section{Dataset Details}
\label{app:dataset_details}

\subsection{Text-Video Retrieval}

We provide dataset-specific details on preprocessing, text construction, and split statistics for the text-video retrieval benchmarks used in our experiments.
The resulting statistics are summarized in Table~\ref{tab:dataset_statistics}.

\paragraph{DiDeMo.}
Distinct Describable Moments (DiDeMo)~\citep{hendricks2017localizing} contains short videos with multiple moment-level descriptions.
As in prior text-video retrieval studies~\citep{liu2019use,bain2021frozen,luo2022clip4clip,ma2022xclip}, we concatenate all captions associated with the same video and formulate the task as paragraph-video retrieval.
In our processed split, the training set contains 8,381 video-text pairs and the test set contains 1,003 video-text pairs.
Each video has 3.9 captions on average in the training split.
The captions are merged into a single text input using whitespace concatenation.

\paragraph{ActivityNet Captions.}
ActivityNet Captions~\citep{krishna2017dense} contains YouTube videos annotated with dense event descriptions.
We use the \texttt{val\_1} split for evaluation and construct paragraph-level text queries by concatenating all captions associated with the same video, consistent with standard text-video retrieval protocols~\citep{liu2019use,bain2021frozen,luo2022clip4clip,ma2022xclip}.
The training set contains 10,009 video-text pairs, and the test set contains 4,917 video-text pairs.
Each video has 3.7 captions on average in the training split.
Compared with DiDeMo, ActivityNet contains longer videos and multiple event-level captions per video.

\paragraph{LSMDC.}
The Large Scale Movie Description Challenge (LSMDC)~\citep{rohrbach2017movie} consists of short movie clips paired with captions from movie scripts or descriptive video services.
We follow the standard text-video retrieval setup used in previous works~\citep{liu2019use,bain2021frozen,luo2022clip4clip,ma2022xclip} and evaluate the model on the 1,000-sample test split.
Our processed training set contains 101,055 samples, and the test split contains 1,000 samples.

\paragraph{MSR-VTT.}
MSR-VTT~\citep{xu2016msrvtt} contains 10K video clips from diverse categories, where each training video is annotated with multiple captions.
We adopt the standard 1K-A test protocol~\citep{yu2018joint,liu2019use}, using 9,000 training videos with 20 captions per video, resulting in 180,000 training text-video pairs.
The test split contains 1,000 videos.
In our dataloader, each caption is treated as an individual text input during training, while the video feature is shared across captions from the same video.

\begin{table*}[t]
\centering
\small
\setlength{\tabcolsep}{7pt}
\renewcommand{\arraystretch}{1.15}
\caption{
Dataset statistics used in our text-video retrieval experiments.
DiDeMo and ActivityNet use paragraph-level text inputs by concatenating multiple captions of the same video, while LSMDC and MSR-VTT use one caption per training sample.
}
\label{tab:dataset_statistics}
\resizebox{0.85\textwidth}{!}{
\begin{tabular}{lcccc}
\toprule
\textbf{Dataset}
& \textbf{Train Samples}
& \textbf{Test Samples}
& \textbf{Caption Format}
& \textbf{Captions / Video} \\
\midrule
DiDeMo      & 8,381   & 1,003 & list of captions & 3.9 \\
ActivityNet & 10,009  & 4,917 & list of captions & 3.7 \\
LSMDC       & 101,055 & 1,000 & single caption   & 1.0 \\
MSR-VTT     & 180,000 & 1,000 & single caption   & 20.0 \\
\bottomrule
\end{tabular}
}
\end{table*}
\label{app:text_video_retrieval_datasets}

\begin{table*}[t]
\centering
\small
\setlength{\tabcolsep}{7pt}
\renewcommand{\arraystretch}{1.12}
\caption{Training settings for each text-video retrieval dataset. Shared settings are applied across all benchmarks, while dataset-specific settings such as the number of slots, total epochs, learning rate, and effective batch size are reported separately.
}
\label{tab:supp_training_settings}
\begin{adjustbox}{max width=\textwidth}
\begin{tabular}{l|cccc}
\toprule
& \textbf{DiDeMo} & \textbf{ActivityNet} & \textbf{LSMDC} & \textbf{MSR-VTT} \\
\midrule
Optimizer & \multicolumn{4}{c}{AdamW} \\
AdamW betas $(\beta_1,\beta_2)$ & \multicolumn{4}{c}{$(0.9, 0.95)$} \\
Weight decay & \multicolumn{4}{c}{0.05} \\
Warmup epochs & \multicolumn{4}{c}{1} \\
Input frames & \multicolumn{4}{c}{16} \\
$\lambda_{\mathrm{div}}$ & \multicolumn{4}{c}{0.1} \\
$\lambda_{\mathrm{hn}}$ & \multicolumn{4}{c}{0.05} \\
Hard-negative margin $\delta$ & \multicolumn{4}{c}{0.04} \\
\midrule
Number of slots $M$ & 4 & 4 & 4 & 3 \\
Total epochs & 5 & 5 & 3 & 3 \\
Learning rate & 2e-4 & 2e-4 & 2e-4 & 1e-4 \\
Effective batch size & 320 & 320 & 320 & 512 \\
\bottomrule
\end{tabular}
\end{adjustbox}
\end{table*}

\section{Implementation Details}
\label{sec:supp_implementation}

\paragraph{Model architecture and trainable parameters.}
MARS is built on VideoChat-Flash-Qwen2-7B~\cite{videochat_flash}.
This MLLM combines a UMT-L vision encoder~\cite{li2023umt}, a linear multimodal projection layer, and a Qwen2 language model~\cite{qwen2}.
The language model contains 28 decoder layers.
Following the prompt design in Sec.~\ref{sec:method}, we insert adaptive representation tokens into the response field for both video and text inputs.
We use the hidden states of these tokens from all decoder layers for layer fusion.
The number of slots is set to four for DiDeMo, ActivityNet, and LSMDC, and three for MSR-VTT.

We fine-tune the model in a parameter-efficient manner.
LoRA~\cite{lora} is applied to the attention and MLP projection modules of the language model, as well as to the multimodal projection layer.
We set the LoRA rank to 128, the LoRA scaling factor to 256, and the dropout rate to 0.05.
The trainable parameters consist of the LoRA parameters, the adaptive representation tokens, and the slot-wise layer-fusion weights.
All other pretrained parameters are frozen during training.

\paragraph{Training configuration.}
We sample 16 frames from each video in all experiments.
The learning rate is linearly warmed up during the first epoch.
To reduce activation memory, we use gradient checkpointing~\cite{gradient_checkpointing}, which allows larger effective batch sizes for contrastive learning.
All models are trained on two NVIDIA A100 40GB GPUs.
Dataset-specific hyperparameters are reported in Table~\ref{tab:supp_training_settings}.

\section{Slot Aggregation Analysis}
\label{app:slot_aggregation}
MARS computes the final retrieval score by aggregating the similarities of corresponding text-video slots.
In the main method, we use uniform aggregation, where each slot contributes equally to the final score.
This design avoids introducing additional slot-weighting parameters and preserves the contribution of all adaptive representation slots.
Table~\ref{tab:ablation_aggregation} compares uniform aggregation with two alternative strategies.
The learnable strategy assigns trainable weights to different slots, while MaxSim selects the strongest slot-level similarity.
Uniform aggregation achieves the best average performance across the eight T2V/V2T settings, with an average R@1 of 65.7, compared with 65.4 for learnable aggregation and 65.1 for MaxSim.
These results suggest that the slots encode complementary retrieval cues, and that a simple average combines them more robustly than either learned weighting or maximum-based selection.

\begin{table}[t]
\centering
\scriptsize
\setlength{\tabcolsep}{3.0pt}
\renewcommand{\arraystretch}{1.08}
\caption{
Ablation on slot aggregation strategies.
}
\label{tab:ablation_aggregation}

\begin{adjustbox}{max width=\columnwidth}
\begin{tabular}{l|cc|cc|cc|cc|c}
\toprule
\multirow{2}{*}{Aggregation}
& \multicolumn{2}{c|}{DiDeMo}
& \multicolumn{2}{c|}{ActivityNet}
& \multicolumn{2}{c|}{LSMDC}
& \multicolumn{2}{c|}{MSR-VTT}
& \multirow{2}{*}{Avg.} \\
\cmidrule(lr){2-3}
\cmidrule(lr){4-5}
\cmidrule(lr){6-7}
\cmidrule(lr){8-9}
& T2V & V2T
& T2V & V2T
& T2V & V2T
& T2V & V2T
& \\
\midrule

Learnable
& 78.6 & 74.4
& 74.8 & \textbf{73.4}
& 50.8 & 50.2
& 61.1 & 59.5
& 65.4 \\

MaxSim
& 78.2 & 73.9
& 74.6 & 72.7
& 50.4 & \textbf{50.3}
& 60.8 & \textbf{59.7}
& 65.1 \\

\rowcolor{MarsCyanA}
\textbf{Uniform}
& \textbf{79.7} & \textbf{75.7}
& \textbf{75.7} & 73.2
& \textbf{51.0} & 50.0
& \textbf{61.6} & 59.0
& \textbf{65.7} \\

\bottomrule
\end{tabular}
\end{adjustbox}
\end{table}

\begin{figure*}[t]
\centering
\includegraphics[width=0.97\textwidth]{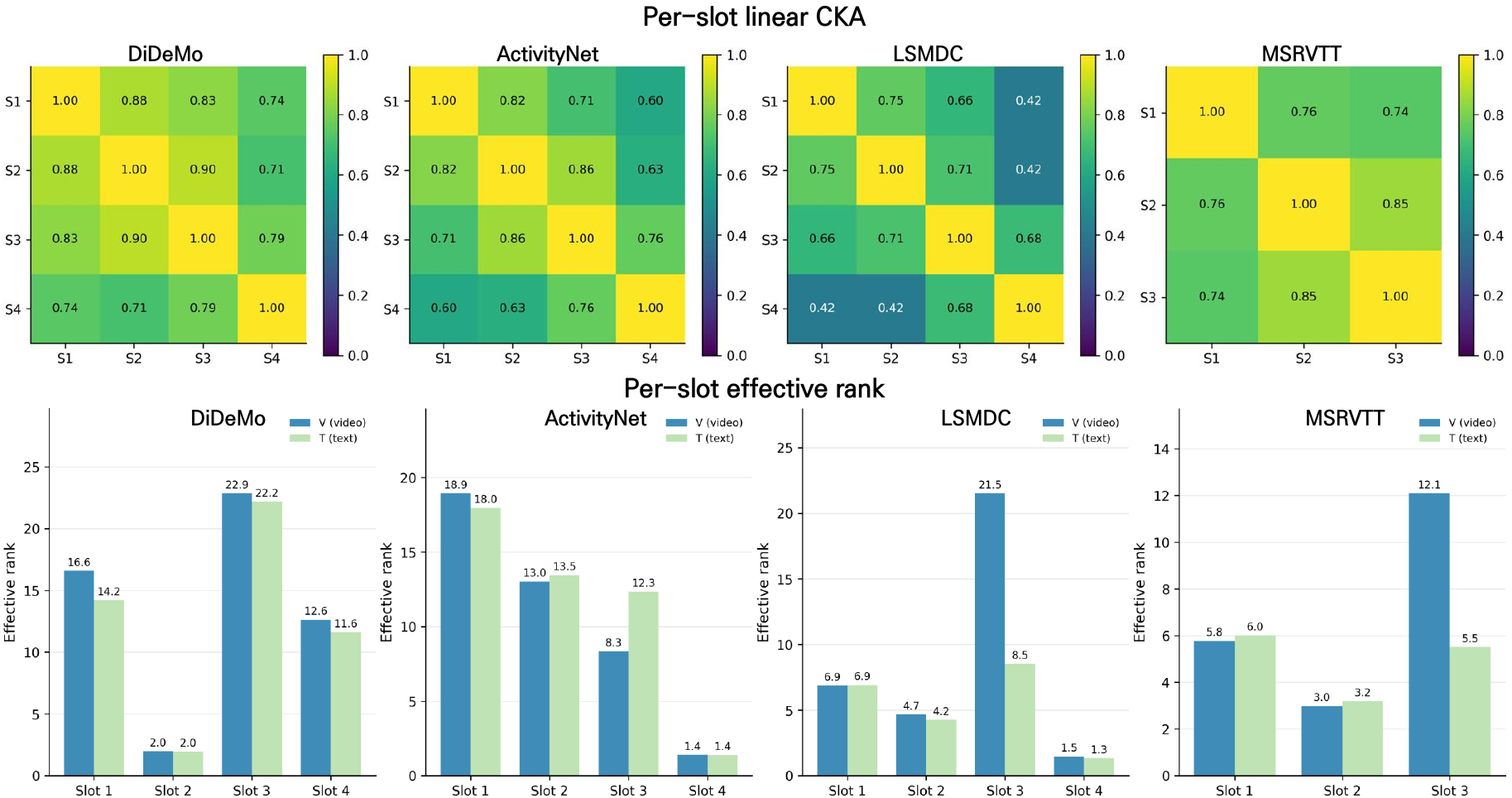}
\vspace{-0.4em}
\caption{
Additional geometry diagnostics of adaptive representation slots.
The first panel reports linear CKA between slot embedding matrices, which measures sample-level structural similarity between slots.
The second panel reports the effective rank of each slot embedding matrix, where a higher effective rank indicates that the slot uses a broader set of embedding directions.
}
\label{fig:app_slot_geometry}
\end{figure*}

\section{Additional Analysis of Adaptive Representation Slots}
\label{app:slot_analysis}

We provide additional analyses of adaptive representation slots.
These results complement the main analysis in Sec.~\ref{sec:analysis} by examining the geometry of slot embeddings from additional perspectives.
All analyses follow the same notation as Sec.~\ref{sec:method}, where each input is represented by multiple adaptive representation slots and the final retrieval score is computed by uniform aggregation of the corresponding slot-wise similarities.

\subsection{Slot Geometry Diagnostics}
\label{app:slot_geometry}

We further examine the geometry of adaptive representation slots.
These diagnostics are not used as retrieval metrics, but provide additional evidence about how the slots relate to one another in the embedding space.
Since pairwise cosine similarity is discussed in Sec.~\ref{sec:analysis}, we focus here on two complementary diagnostics: linear CKA and effective rank.

Pairwise cosine similarity compares individual vector directions, but does not fully describe whether two slots encode similar sample-level structures.
We therefore compute linear CKA between slot embedding matrices.
As shown in the first panel of Figure~\ref{fig:app_slot_geometry}, the structural similarity between slots varies across benchmarks.
This suggests that slot distinctness can depend on dataset characteristics.
Together with the pairwise cosine analysis in Sec.~\ref{sec:analysis}, the CKA analysis provides a complementary view of how adaptive representation slots differ beyond their vector directions.

We also measure the effective rank of each slot embedding matrix.
The effective rank is computed from the entropy of the normalized squared singular values.
It reflects how broadly a slot uses the available embedding dimensions.
The second panel of Figure~\ref{fig:app_slot_geometry} shows that different slots can have different effective ranks.
This indicates that adaptive representation slots do not necessarily use the embedding space in the same way.
Combined with the retrieval-side analyses in Sec.~\ref{sec:analysis}, these geometry diagnostics suggest that the slots form distinct representation patterns.

\definecolor{DirectBlock}{RGB}{235,240,248}
\definecolor{ScoringBlock}{RGB}{237,245,240}
\definecolor{MarsCyanLight}{RGB}{224,247,250}
\definecolor{MarsGreenLight}{RGB}{225,247,235}

\end{document}